\documentclass{article}
\usepackage{graphicx}
\usepackage[utf8]{inputenc}
\usepackage{amssymb,amsmath,array}
\usepackage{setspace} 

\usepackage[style=unsrt,url=false,isbn=false,doi=false,style=numeric-comp,giveninits,backend=biber,sorting=none]{biblatex}

\usepackage{hyperref}
\usepackage{url} 
\usepackage{epstopdf}
\usepackage{upgreek}

\usepackage[dvipsnames]{xcolor}
\usepackage[footnotesize]{caption}

\usepackage[leftcaption]{sidecap}

\newcommand{\ie}{\emph{i.e.}, }
\newcommand{\eg}{\emph{e.g.}, }
\newcommand{\iid}{%
    \ifmmode
        \mathrm{i.i.d.}%
    \else%
        i.i.d.\@\xspace%
    \fi%
}

\newcommand{\ts}{\textstyle}
\newcommand{\bs}{\boldsymbol}
\newcommand{\cl}{\mathcal}
\newcommand{\bb}{\mathbb}
\newcommand{\wh}{\widehat}
\renewcommand{\Vec}[1]{\bs{#1}} 
\newcommand{\distiid}{\sim_\iid} 
\newcommand{\expec}[1]{\mathop{{}\mathbb{E}}_{#1}} 

\DeclareMathOperator{\diag}{diag}
\newcommand{\im}{\mathrm{i}\mkern1mu} 

\newcommand{\sigsp}{\Sigma}

\usepackage{ifthen}
\newif\ifcomment
\commenttrue

\ifcomment
\newcommand{\LJc}[1]{\textcolor{blue}{\textbf{\scriptsize [LJ: #1]}}}
\newcommand{\VSc}[1]{\textcolor{SeaGreen}{\textbf{\scriptsize [VS: #1]}}}
\newcommand{\todo}[1]{\textcolor{red}{\textbf{\scriptsize [TODO: #1]}}}
\newcommand{\removable}[1]{\textcolor{green}{#1}}
\else
\newcommand{\LJc}[1]{}
\newcommand{\VSc}[1]{}
\newcommand{\todo}[1]{}
\newcommand{\removable}[1]{}
\fi

\newcommand{\delparvs}{\vspace{-4mm}}

\begin{document}
\title{Compressive Learning of Generative Networks}

\date{}
\author{Vincent Schellekens and Laurent Jacques\thanks{ISPGroup, ICTEAM, UCLouvain, Louvain-La-Neuve - Belgium. VS and LJ are funded by the Belgian National Science Foundation (F.R.S.-FNRS).}
%
\vspace{.3cm}\\
%
}

\maketitle

\ifcomment
\fi

\begin{abstract}
Generative networks implicitly approximate complex densities from their sampling with impressive accuracy. However, because of the enormous scale of modern datasets, this training process is often computationally expensive. We cast generative network training into the recent framework of \emph{compressive learning}: we reduce the computational burden of large-scale datasets by first harshly compressing them in a single pass as a single sketch vector. We then propose a cost function, which approximates the Maximum Mean Discrepancy metric, but requires \emph{only} this sketch, which makes it time- and memory-efficient to optimize. 
\end{abstract}

\section{Introduction}
These last few years, data-driven methods took over the state-of-the-art in a staggering amount of research and engineering applications. This success owes to a combination of two factors: machine learning models that combine expressive power and good generalization properties (\eg deep neural networks), and unprecedented availability of training data in enormous quantities.

Among such models, \emph{generative networks} (GNs) received a significant amount of interest for their ability to embed data-driven priors in general applications, \eg for solving inverse problems such as super-resolution, deconvolution, inpainting, or compressive sensing to name a few~\cite{bora2017compressed,mardani2017deep,rick2017one,Lucas_2018}. As explained in Sec.~\ref{sec:background}, GNs are deep neural networks (DNNs) trained to generate samples that mimic those available in a given dataset. By minimizing some well-crafted cost-function at the training, these networks implicitly learn the probability distribution synthesizing this dataset; passing randomly generated low-dimensional inputs through to the GN then generates new high-dimensional samples.

In generative \emph{adversarial} networks (GANs) this cost is dictated by a discriminator network that classifies real (training) and fake (generated) examples, the generative and the discriminator networks being learned simultaneously in a two-player zero-sum game~\cite{goodfellow2014generative}. While GANs are the golden standard, achieving the state-of-the-art for a wide variety of tasks, they are notoriously hard to learn due to the need to balance carefully the training of the two networks. 

MMD-GNs minimize the simpler Maximum Mean Discrepancy (MMD) cost function~\cite{li2015generative,dziugaite2015training}, \ie a ``kernelized'' distance measuring the similarity of generated and real samples. Although training MMD-GNs is conceptually simpler than GANs --- we can resort to simple gradient descent-based solvers (\eg SGD) --- its computational complexity scales poorly with large-scale datasets: each iteration necessitates numerous (typically of the order of thousands) accesses to the whole dataset. This severely limits the practical use of MMD-GNs~\cite{arjovsky2017wasserstein}. 

Indeed, modern machine learning models such as GN are typically learned from numerous (\eg several million) training examples. Aggregating, storing, and learning from such large-scale datasets is a serious challenge, as the required communication, memory, and processing resources inflate accordingly. In compressive learning (CL), larger datasets can be exploited without demanding more computational resources.
The data is first harshly compressed to a single vector called the \emph{sketch}, a process done in a single, easily parallelizable pass over the dataset~\cite{gribonval2017compressiveStatisticalLearning}. The actual learning is then performed from the sketch only, which acts as a light proxy for the whole dataset statistics. However, CL has for now been limited to ``simple'' models explicitly parametrized by a handful of parameters, such as k-means clustering, Gaussian mixture modeling or PCA~\cite{gribonval2017compressiveStatisticalLearning}.

\begin{figure}
	\centering
	\includegraphics[width=0.9\linewidth]{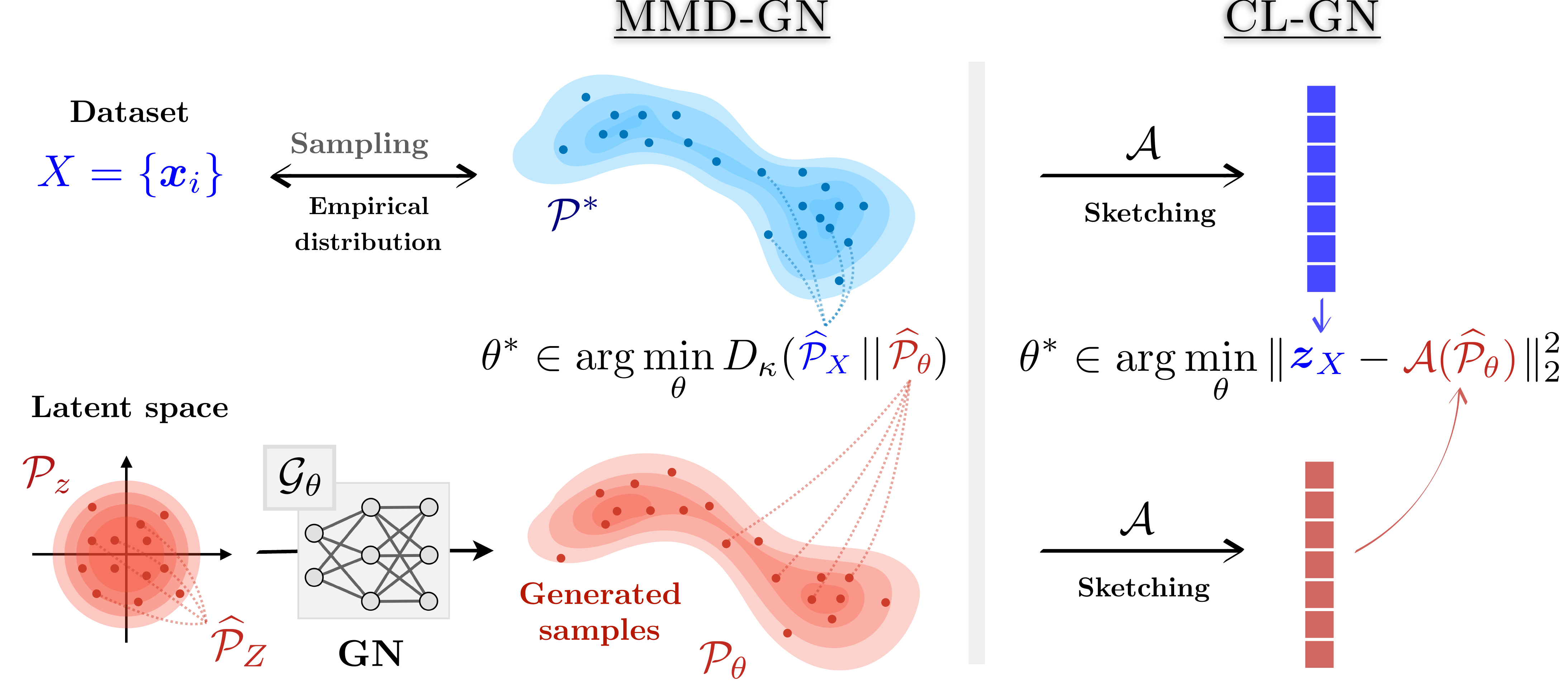}
	\caption{General overview of our approach. The moment matching of MMD-GNs is replaced by sketching both $X$ and the sampling $\widehat{\cl P}_\theta$. This compressive learning approach of GNs (or CL-GNs) is allowed by relating the RFF frequency distribution $\Lambda$ to the MMD kernel $\kappa$.} 
	\label{fig:main}
\end{figure}

This work proposes and assesses the potential of sketching to ``compressively learn'' deep generative networks (MMD-GNs) with greatly reduced computational cost (see Fig.~\ref{fig:main}). By defining a cost function and practical learning scheme, our approach serves as a prototype for compressively learning general generative models from sketches. The effectiveness of this scheme is tested on toy examples.

\section{Background, related work and notations}
\label{sec:background}
To fix the ideas, given some space $\sigsp \subset \bb R^d$, we assimilate any dataset $X = \{\bs x_i\}_{i=1}^n \subset \sigsp$ with $n$ samples to a discrete probability measure $\wh{\cl P}_{X}$, \ie an empirical estimate for the \emph{probability distribution} $\cl P^*$ generating $X$. Said differently, $\Vec{x}_i \sim_\iid \cl P^*$ and $\wh{\cl P}_{X} := \frac{1}{|X|}\sum_{i=1}^n \delta_{\Vec{x}_i}$, where $\delta_{\Vec{c}}$ is the Dirac measure at $\Vec{c} \in \sigsp$. 

\delparvs
\paragraph{2.1.~Compressive Learning:} In CL, massive datasets are first efficiently (in one parallelizable pass) compressed into a single \emph{sketch} vector of moderate size. The required parameters are then extracted from this sketch, using limited computational resources compared to usual algorithms that operate on the full dataset~\cite{gribonval2017compressiveStatisticalLearning}.

The sketch operator $\ts \cl A(\cl P) := \frac{1}{\sqrt{m}} \left[ \, \expec{\Vec{x} \sim \cl P}  \exp \left(\im \Vec{\omega}_j^T\Vec{x}\right) \right]_{j=1}^m$ realizes an embedding of any (infinite dimensional) probability measure $\cl P$ into the low-dimensional domain $\bb C^m$. This sketching amounts to taking the expectation of the random Fourier features (RFF)~\cite{Rahimi2008RFF} $\Phi(\Vec{x}) := \frac{1}{\sqrt{m}} \exp(\im\bs\Omega^T\Vec{x})$ of $\bs x \sim \cl P$, with $\Omega := (\Vec{\omega}_1,\,\cdots, \Vec{\omega}_m) \in \bb R^{d \times m}$. For large values of $n$, we expect that \begin{equation}
\label{eq:sketch}
\ts \cl A(\cl P^*)\ \approx\ \Vec{z}_{X} := \cl A(\wh{\cl P}_{X}) = \frac{1}{n} \sum_{i=1}^n \Phi(\Vec{x}_i) \in \bb C^m,
\end{equation}
where $z_X$ is the \emph{sketch of the dataset} $X$. This sketch, which has a \emph{constant size} $m$ whatever the cardinality of $X$, thus embeds $\wh{\cl P}_{X}$ by empirically averaging (or \emph{pooling}) all RFF vectors $\Phi(\Vec{x}_i)$. We still need to specify the RFF projection matrix $\bs \Omega$; it is randomly generated by drawing $m$ ``frequencies'' $\Vec{\omega}_j \distiid \Lambda$. In other words, $\cl A(\cl P)$ corresponds here to a random sampling (according to the law~$\Lambda$) of the characteristic function of $\cl P$ (\ie its Fourier transform). By Bochner's theorem~\cite{Rudin1962bochnerBook},~$\Lambda$ is related to some shift-invariant \emph{kernel} $\kappa(\Vec{x},\Vec{y}) = K(\Vec{x}-\Vec{y})$ by the (inverse) Fourier transform: $K(\Vec{u}) = \expec{\Vec{\omega} \sim \Lambda} e^{\im \Vec{\omega}^T\Vec{u}} =: \cl F^{-1}[\Lambda](\bs u)$.

CL aims at learning, from only the sketch $\Vec{z}_X$, an approximation $\cl P_{\theta}$ for the density $\cl P^*$, parametrized by $\theta \in \Theta$. For example, $\theta$ collects the position of the $K$ centroids for compressive $K$-means, and the weights, centers and covariances of different Gaussians for compressive Gaussian mixtures fitting. This is achieved by solving the following density fitting (``sketch matching'') problem:
\begin{equation}
\label{eq:sketch_matching}
\theta^* \in \arg \min_{\theta} \| \Vec{z}_X - \cl A(\cl P_{\theta}) \|^2_2 \quad \text{s.t.} \quad \theta \in \Theta.
\end{equation}

For large values of $m$, the cost in \eqref{eq:sketch_matching} estimates a metric $D_\kappa$ between~$\wh{\cl P}_X$ and~$\cl P_\theta$, called the Maximum Mean Discrepancy (MMD)~\cite{gretton2012kernelTwoSample}, that is kernelized by $\kappa$, \ie writing $\kappa(\cl P, \cl Q) := \bb E_{\Vec{x} \sim \cl P, \Vec{y} \sim \cl Q }\,\kappa(\Vec{x},\Vec{y})$, the MMD reads
\begin{equation}
\label{eq:MMD}
	D_{\kappa}^2(\cl P,\cl Q) := \kappa(\cl P,\cl P) + \kappa(\cl Q,\cl Q) - 2\kappa(\cl P,\cl Q).
\end{equation}
Using Bochner's theorem, we can indeed rewrite~\eqref{eq:MMD} as
\begin{multline}
\label{eq:sketch_MMD}
	\ts \| \cl A(\wh{\cl P}_{X}) - \cl A(\cl P_{\theta}) \|^2_2  = \frac{1}{m} \sum_{j = 1}^m \big| \expec{\Vec{x}\sim\wh{\cl P}_{X}} e^{\im\Vec{\omega}_j^T\Vec{x} } - \expec{\Vec{y}\sim\cl P_{\theta}} e^{\im\Vec{\omega}_j^T\Vec{y} } \big|^2\\
	\ts \simeq \expec{\Vec{\omega} \sim \Lambda} \big| \expec{\Vec{x}\sim\wh{\cl P}_{X}} e^{\im\Vec{\omega}^T\Vec{x} } - \expec{\Vec{y}\sim\cl P_{\theta}} e^{\im\Vec{\omega}^T\Vec{y} } \big|^2 = D_{\kappa}^2(\wh{\cl P}_{X},\cl P_{\theta}).
\end{multline} 
Provided $\Lambda$ is supported on $\bb R^d$, $D_\kappa(\cl P, \cl Q) = 0$ if and only if $\cl P = \cl Q$~\cite{Sriperumbudur2010hilbertEmbedding}. Thus, minimizing \eqref{eq:sketch_matching} accurately estimates $\wh{\cl P}_{X}$ from $\cl P_{\theta^*}$ if $m$ is large compared to the complexity of the model; \eg in compressive K-means, CL requires experimentally $m = O(Kd)$ to learn the centroids of $K$ clusters in $\bb R^d$. 

The non-convex sketch matching problem \eqref{eq:sketch_matching} is generally solved with greedy heuristics (\eg CL-OMPR~\cite{keriven2016GMMestimation}). As they require a closed-form expression of $\cl A(\cl P_{\theta})$ and the Jacobian $\nabla_{\theta} \cl A(\cl P_{\theta})$, CL has so far be limited to cases where $\cl P_{\theta}$ is explicitly available and easy to manipulate.

\delparvs
\paragraph{2.2.~Generative networks:} To generates realistic data samples, a GN $\cl G_{\theta^*} : \Sigma_z \mapsto \sigsp$ (\ie a DNN) with weights $\theta^* \in \bb R^{d_{\theta}}$ is trained as follows. Given $\theta \in \Theta$, we compute the empirical distribution 
$\wh{\cl P}_{\theta} := \cl G_{\theta}(\wh{\cl P}_Z) = \frac{1}{n'} \sum_{i = 1}^{n'} \cl G_{\theta}(\Vec{z}_i)$ of $n'$ inputs $Z = \{\Vec{z}_{i}\}_{{i} = 1}^{n'}$ randomly drawn in a low-dimensional \emph{latent space} $\Sigma_z \subset \mathbb{R}^p$ from a simple distribution $\cl P_z$, \eg $\Vec{z}_i \distiid \cl N(\Vec{0},\Vec{I}_p)$. By design, $\wh{\cl P}_\theta$ is related to sampling the \emph{pushforward} distribution of $\cl P_z$ by $\cl G_{\theta}$. The parameter $\theta^*$ is then set such that $\wh{\cl P}_{\theta^*} \approx \wh{\cl P}_{X}$. While several divergences have been proposed to quantify this objective, we focus here on minimizing the MMD metric $D_{\kappa}^2(\wh{\cl P}_{X},\wh{\cl P}_{\theta})$~\cite{li2015generative,dziugaite2015training}. Using \eqref{eq:MMD} and discarding constant terms, we get the MMD-GNs fitting problem: 
\begin{equation}
\label{eq:MMD-GN}
\ts \theta^* = \arg\min_{\theta}\  \sum_{\bs z_i, \bs z_j \in Z} \kappa(\cl G_{\theta}(\bs z_i), \cl G_{\theta}(\bs z_j)) - 2\sum_{\bs x_i \in X, \bs z_j \in Z} \kappa(\bs x_i , \cl G_{\theta}(\bs z_j)).
\end{equation}

Li et al. called this approach generative \emph{moment matching} networks, as minimizing~\eqref{eq:MMD} amounts to matching all the (infinite) moments of $\cl P$ and $\cl Q$ thanks to the space kernelization yielded by $\kappa$~\cite{Hall2005generalizedMethodMoments}~(see Fig.~\ref{fig:main}).

If $\kappa$ is differentiable, gradient descent-based methods can be used to solve~\eqref{eq:MMD-GN}, using back-propagation to compute the gradients of $\cl G_{\theta}$. However, for $n$ true samples and $n'$ generated samples (or a batch-size), each evaluation of $D_{\kappa}^2(\wh{\cl P}_{X},\wh{\cl P}_{\theta})$ (and its gradient) requires $\cl O(nn'+n'^2)$ computations. Training MMD-GNs, while conceptually simpler than training GANs, is much slower due to all the pairwise evaluations of the kernel required at each iteration --- especially for modern large-size datasets.

\section{Compressive Learning of Generative Networks}
\label{sec:contrib}

In this work, given a dataset $X$, we propose to learn a generative network $\cl G_{\theta}$ using \emph{only} the sketch $\Vec{z}_{X} = \cl A(\wh{\cl P}_X)$ defined in \eqref{eq:sketch} (see Fig.~\ref{fig:main}). For this, given $n'$ samples $Z = \{\Vec{z}_i \sim \cl P_z\}_{i=1}^{n'}$, we solve a \emph{generative network sketch matching problem} that selects $\theta^* = \arg\min_{\theta} \cl L(\theta;\Vec{z}_{X})$ with 
\begin{equation}
\label{eq:proposedcost}
    \ts \cl L(\theta;\Vec{z}_{X}) := \big\|\cl A(\wh{\cl P}_{X}) - \cl A\big(\cl G_{\theta}(\wh{\cl P}_{Z})\big)\big\|_2^2 = \big\|\Vec{z}_{X} - \frac{1}{n'} \sum_{i=1}^{n'} \Phi\big(\cl G_{\theta}(\Vec{z}_i))\big\|_2^2.
\end{equation}

From~\eqref{eq:sketch_MMD}, we reach $\cl L(\theta;\Vec{z}_{X}) \simeq D_{\kappa}(\wh{\cl P}_X \, || \, \cl G_{\theta}(\wh{\cl P}_Z))$ for large values of $m$, as established from the link relating $\kappa$ and $\Lambda$. Compared to the exact MMD in~\eqref{eq:MMD-GN}, $\cl L(\theta;\Vec{z}_{X})$ is, however, much easier to optimize. Once the dataset sketch $\Vec{z}_X$ has been pre-computed (in one single pass over $X$, possibly in parallel), we only need to compute $\cl A(\cl G_{\theta}(\wh{\cl P}_{Z}))$ (\ie by computing $n'$ contributions $\bs z_i \to \Phi(\cl G_{\theta}(\Vec{z}_i))$ by feed-forward, before averaging them) to compute the Euclidean distance between both quantities. In short, we access $X$ only once then discard it, and evaluating the cost has complexity $\cl O(n')$, \ie  much smaller than $\cl O(nn' + n'^2)$, the complexity of the exact MMD~\eqref{eq:MMD-GN} (see Sec.~2.2). 

Equally importantly, the gradient $\nabla_{\theta} \cl L(\theta;\Vec{z}_{X})$ is easily computed. With the residual $\Vec{r} := \Vec{z}_{X} - \frac{1}{n'} \sum_{i=1}^{n'} \Phi (\cl G_{\theta}(\Vec{z}_i))$ and $\Vec{r}^H$ its conjugate transpose, 
\begin{equation}
\label{eq:grad}
\ts \nabla_{\theta} \wh{\cl L}(\theta;\Vec{z}_{X}) = -2 \cdot  \frac{1}{n'} \sum_{i=1}^{n'} \, \Re \big[ \Vec{r}^H \big(  \frac{\partial \Phi(\Vec{u})}{\partial \Vec{u}} \big|_{\Vec{u}=\cl G_{\theta}(\Vec{z}_i)}  \cdot \frac{\partial \cl G_{\theta}(\Vec{z}_i)}{\partial \theta} \big) \big].
\end{equation}
Above, $\frac{\partial \Phi(\Vec{u})}{\partial \Vec{u}} = \frac{\im}{\sqrt{m}} \diag( e^{\im \bs\Omega^T\Vec{u}})\,\bs\Omega$ is the $m \times d$ Jacobian matrix listing the partial derivatives of the $m$ sketch entries with respect to the $d$ dimension of $\Vec{u} \in \sigsp$, which is evaluated at the generated samples $\cl G_{\theta}(\Vec{z}_i)$. The last term $\frac{\partial \cl G_{\theta}(\Vec{z}_i)}{\partial \theta} \in \bb R^{d \times d_{\theta}}$ is computed by the back-propagation algorithm as it contains the derivative of the network output $\cl G_{\theta}(\Vec{z}_i)$ (for $\Vec{z}_i$ \textit{fixed}) with respect to the parameters $\theta \in \bb R^{d_{\theta}}$. Algorithmically, the feature function $\Phi$ amounts to an extra layer on top of the GN, with fixed weights $\Omega$ and activation $t \mapsto \exp(\im t)$. We can then plug those expressions in any gradient-based optimisation solver\footnote{To boost the evaluations of~\eqref{eq:grad}, we can split $Z$ into several minibatches $Z_b$ of size $n_b \ll n'$; \eqref{eq:grad} is then replaced by successive minibatch gradients evaluated on the batches $Z_b$. As reported for MMD-GNs \cite{li2015generative,dziugaite2015training}, this only works for sufficiently large $n_b$, \eg $n_b = 1000$ in Sec.~\ref{sec:experiments}.}. 

We conclude this section by an interesting interpretation of \eqref{eq:proposedcost}. While CL requires closed form expressions for $\cl A(\cl P_{\theta})$ and $\nabla_{\theta}\cl A(\cl P_{\theta})$, our GN formalism actually estimates those quantities by Monte-Carlo sampling, \ie replacing $\cl P_{\theta}$ by $\wh{\cl P}_{\theta}$. This thus opens CL to non-parametric density fitting.

\section{Experiments}
\label{sec:experiments}

\begin{figure}
	\centering 
	\includegraphics[width=0.2867\linewidth]{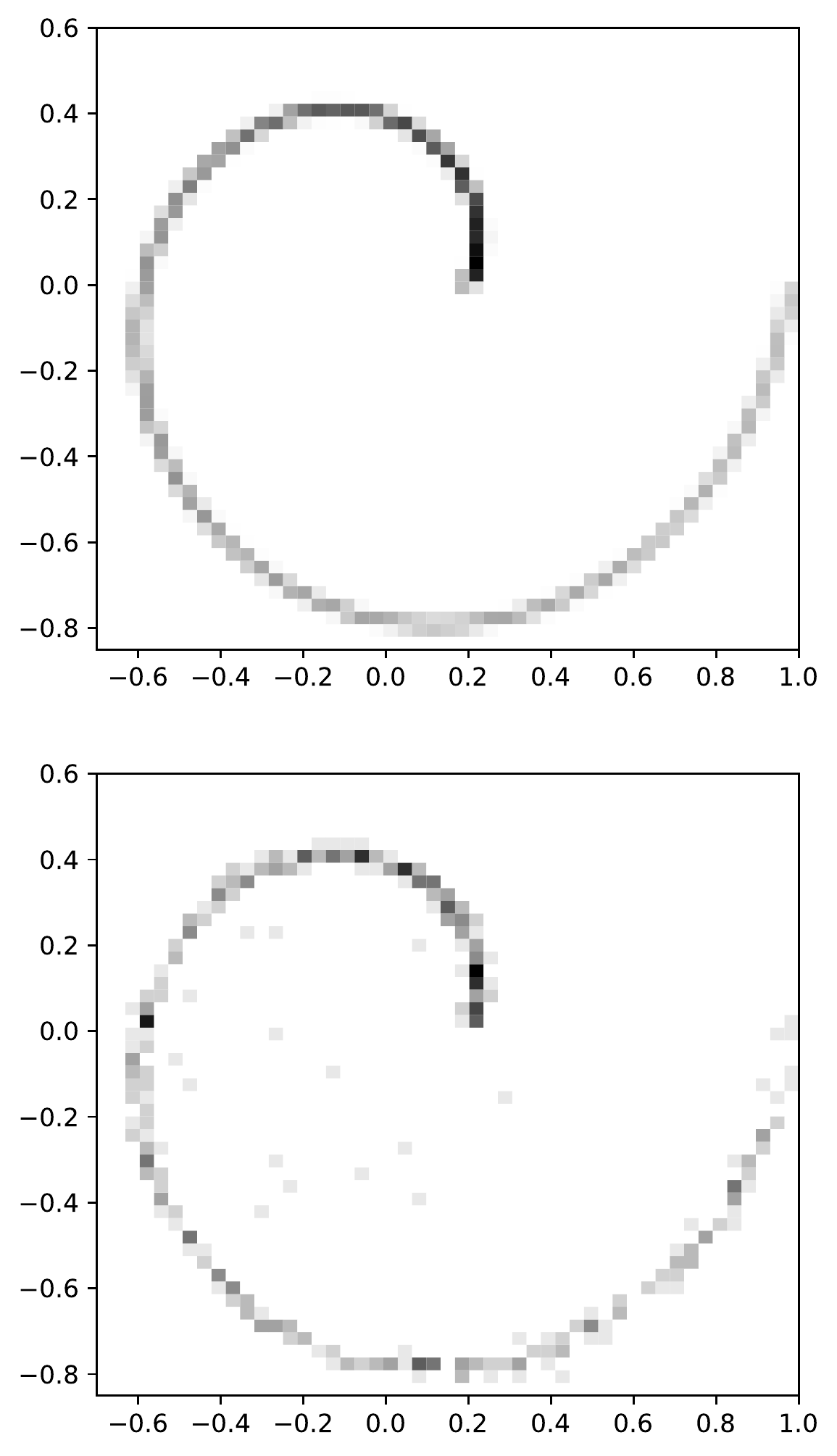}\hspace{-1mm}
	\includegraphics[width=0.2790\linewidth]{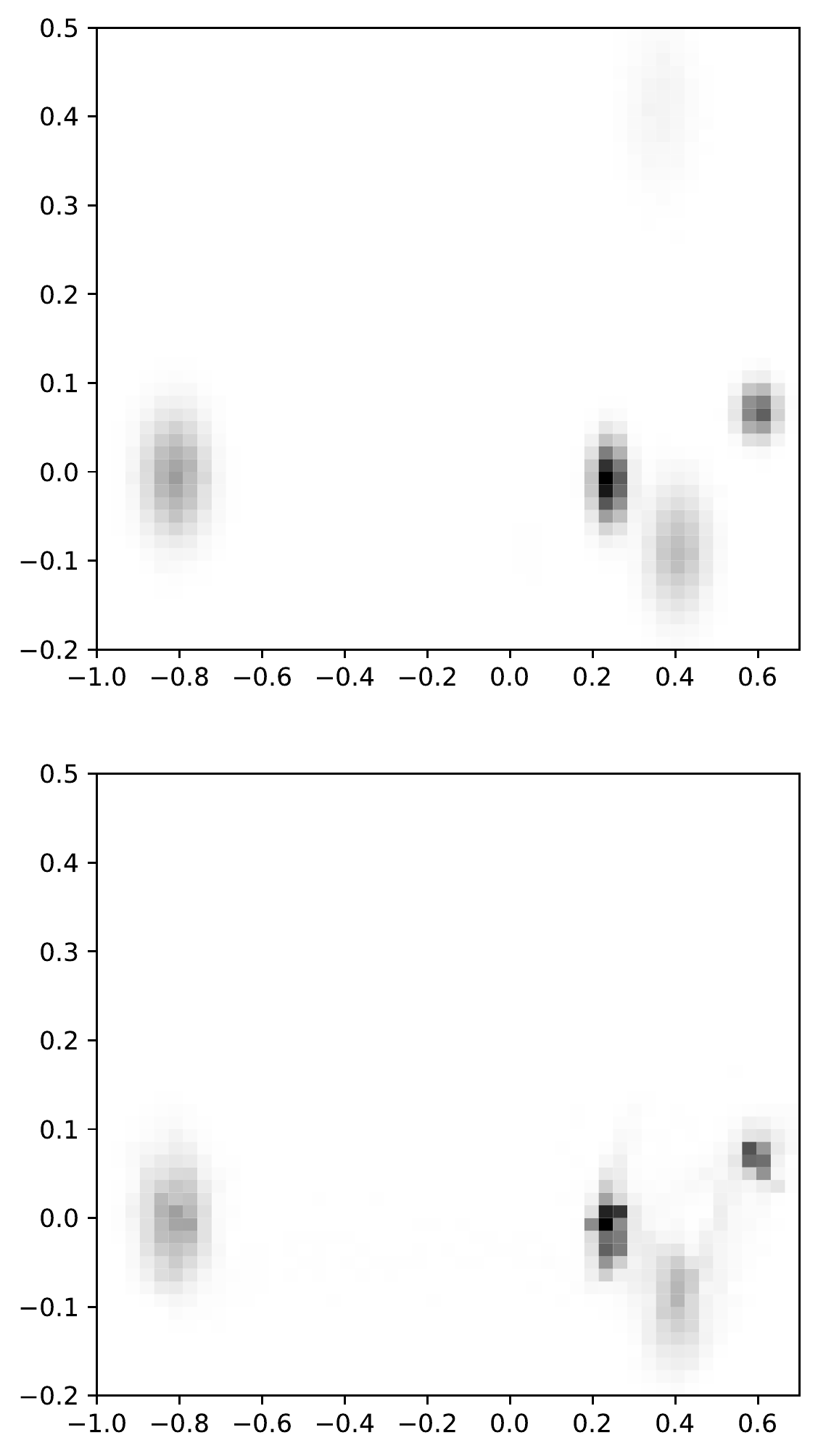}\hspace{-1mm}
	\includegraphics[width=0.2930\linewidth]{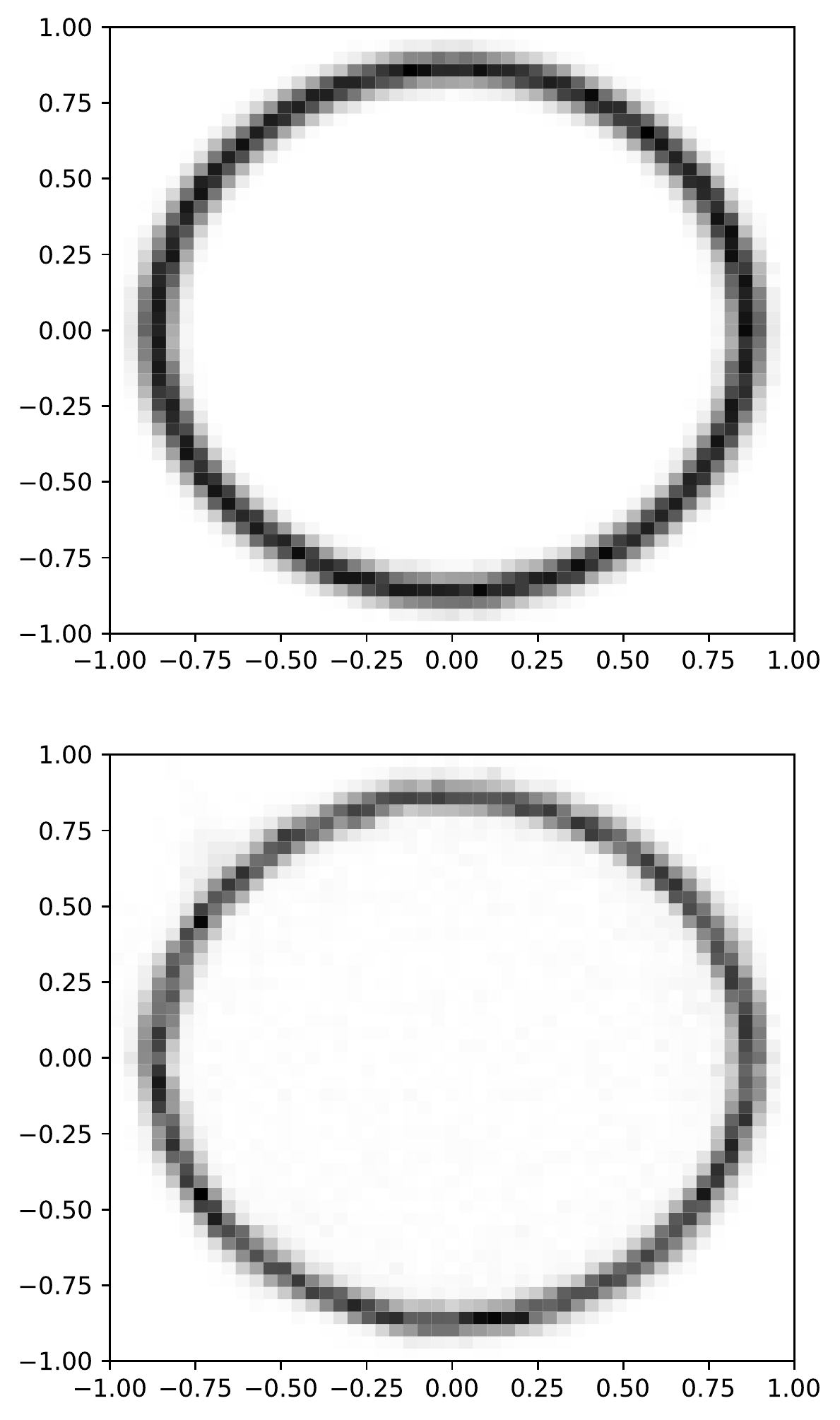}
	
	\caption{Histograms (varying from white to black as the number of samples increases) of considered datasets (top) and $50000$ generated samples, after training from the sketch (bottom).\vspace{-3mm}\label{fig:toys}}
\end{figure}

For this preliminary work, we visually illustrate the effectiveness of minimizing~\eqref{eq:proposedcost} by considering three 2-D synthetic datasets made of $n = 10^5$ samples (see the top row of Fig.~\ref{fig:toys}): \textit{(i)} a 2-D spiral  $\{(r_i,\phi_i)\}_{i=1}^n$, with $\phi_i \sim_\iid \cl U([0,2\pi))$ and $r_i \sim_\iid \frac{\phi_i}{2\pi} + \cl N(0,\sigma_r^2)$, \textit{(ii)} a Gaussian mixture models of $6$ Gaussians, and \textit{(iii)} samples in a circle, \ie $\phi_i \sim \cl U([0,2\pi))$ and $r_i \sim R + \cl N(0, \sigma_r^2)$ for $R$ and $\sigma_r$ fixed. We learn a GN mapping $10-$dimensional random Gaussian vectors to $\bb R^{2}$, passing through seven fully connected hidden layers of $10$ units each, activated by a Leaky ReLU function with slope $0.2$. For this simple illustration, we sketch all datasets to a sketch of size $m = n/10 = 10^4$. We found experimentally from a few trials that setting $\Lambda$ to a folded Gaussian distribution (see~\cite{keriven2016GMMestimation}) of scale $\sigma^2 = 10^{-3}$ is appropriate to draw the $m$ frequencies $\{\Vec{\omega}_j\}_{j=1}^m$. From those sketches, we then trained our generators according to~\eqref{eq:proposedcost}, using the \texttt{keras} framework. We fixed the number of generated samples to $n' = 10^5$, which we split into mini-batches of $n_b = 1000$ samples when computing the gradient. 

Fig.~\ref{fig:toys} compares the densities of generated samples and re-generated samples after the training (from the known densities) through their 2-D histograms. Note that while the datasets are simplistic, we restricted the training time to a few minutes and, except for the frequency distribution, no hyper-parameter tuning was performed. Despite a few outliers and missing probability masses, the visual proximity of the histograms proves the capacity of our method to learn complex 2-D distributions. Our code and further experiments are available at \url{https://github.com/schellekensv/CL-GN}.

\section{Conclusion}
We proposed and tested a method that incorporates compressive learning ideas into generative network training from the Maximum Mean Discrepancy metric. When dealing with large-scale datasets, our approach is potentially orders of magnitude faster than exact MMD-based learning. 

However, to embrace higher-dimensional applications (\eg for image restorations or large scale inverse problems), future works will need to \textit{(i)} devise efficient techniques to adjust the kernel $\kappa$ (\ie the frequency distribution $\Lambda$) to the dataset $X$, and \textit{(ii)} determine theoretically the required sketch size $m$ in function of the dataset distribution $\cl P^*$. Concerning the choice of the kernel, a promising direction consists in tuning its Fourier transform $\Lambda$ directly from a lightweight sketch~\cite{keriven2016GMMestimation}. As for the required sketch size, this problem certainly relates to measuring the ``complexity'' of the true generating density $\cl P^*$, and to the general open question of why over-parametrized deep neural networks generalize so well. 

\singlespacing
\printbibliography

\end{document}